\documentclass{template/ar-1col-S2O}
\usepackage[numbers]{natbib}
\usepackage{url}
\usepackage{comment}
\usepackage{amsmath}
\usepackage[inline]{enumitem}
\usepackage{microtype}

\definecolor{gray}{RGB}{128,128,128}
\definecolor{orange}{RGB}{255,165,0}
\newif\iffinal

\jname{Annual Review of Control, Robotics, and Autonomous Systems}
\jvol{AA}
\jyear{YYYY}
\doi{10.1146/((please add article doi))}

\begin{document}

% Page header
\markboth{Lagomarsino, Merlo et al.}{Programming, Planning, and Allocation in Synergistic HRC}

% \title{A Review of Intuitive Programming, Adaptive Task Planning, and Dynamic Role Allocation
% \vspace{-0.2cm}\\ \vspace{-0.3cm}{\Large 
% for Synergistic Human-Robot Collaboration
% }\vspace{-0.2cm}}

\title{Intuitive Programming, Adaptive Task Planning, and Dynamic Role Allocation in Human-Robot Collaboration}

%Authors, affiliations address.
\author{Marta Lagomarsino$^{1, \dagger}$, Elena Merlo$^{1, 2\dagger}$, Andrea Pupa$^3$, Timo Birr$^4$, Franziska Krebs$^4$, Cristian Secchi$^3$, Tamim Asfour$^4$, and Arash Ajoudani$^1$
\affil{$^1$Human-Robot Interfaces and Interaction (HRI$^2$) lab, Istituto Italiano di Tecnologia (IIT), Genova, Italy.}
\affil{$^2$ Dipartimento di Informatica, Bioingegneria, Robotica e Ingegneria dei Sistemi (DIBRIS), Università degli Studi di Genova, Genova, Italy.}
\affil{$^3$Dipartimento di Scienze e Metodi dell'Ingegneria, Università di Modena e Reggio Emilia (UNIMORE), Reggio Emilia, Italy.}
\affil{$^4$Institute for Anthropomatics and Robotics, Karlsruhe Institute of Technology, Karlsruhe, Germany.}
%\vspace{-0.05cm}
}

%Abstract
\begin{abstract}
%\vspace{-0.3cm}
Remarkable capabilities have been achieved by robotics and AI, mastering complex tasks and environments. Yet, humans often remain passive observers, fascinated but uncertain how to engage. Robots, in turn, cannot reach their full potential in human-populated environments without effectively modeling human states and intentions and adapting their behavior. 
To achieve a synergistic human-robot collaboration (HRC), a continuous information flow should be established: humans must intuitively communicate instructions, share expertise, and express needs.
In parallel, robots must clearly convey their internal state and forthcoming actions to keep users informed, comfortable, and in control. This review identifies and connects key components enabling intuitive information exchange and skill transfer between humans and robots. We examine the full interaction pipeline: from the human-to-robot communication bridge translating multimodal inputs into robot-understandable representations, through adaptive planning and role allocation, to the control layer and feedback mechanisms to close the loop. 
Finally, we highlight trends and promising directions toward more adaptive, accessible HRC.% synergy.
%\vspace{-0.1cm}
\end{abstract}

%Keywords, etc.
\begin{keywords}
%\vspace{-0.2cm}
intuitive programming, adaptive task planning,  dynamic role allocation, human-robot collaboration, multimodal interaction.
%\vspace{-0.2cm}
\end{keywords}
\maketitle

%Table of Contents
\tableofcontents

%%%%%
\section{INTRODUCTION}
\label{sec:introduction}

\begin{figure}[!t]
    \centering
    \includegraphics[width=1.2\textwidth]{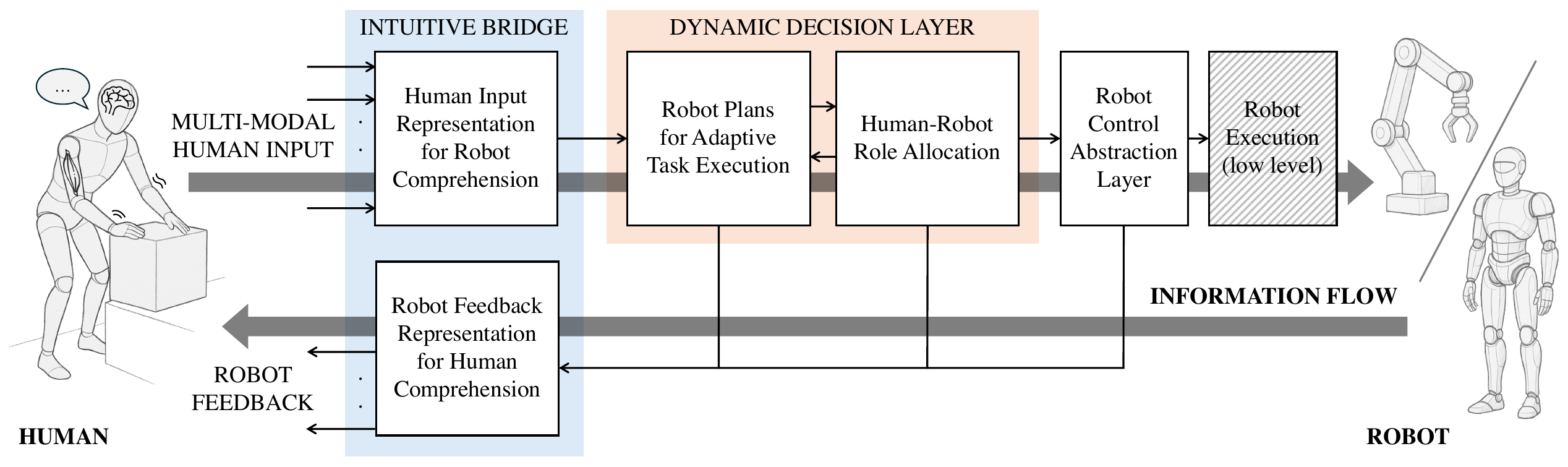}
    \caption{Synergistic HRC systems function as dynamic, bidirectional channels where humans and robots continuously exchange information to enable shared understanding and coordinated action.}
    \label{fig:pipeline}
\end{figure}

The rapid evolution of human-robot collaboration (HRC) promises more flexible, efficient, and intelligent shared workspaces \cite{ajoudani2018progress}. As these systems become increasingly complex and interactive, effective collaboration and true \emph{synergy}, where the combined performance of human and robot exceeds what either could achieve alone, increasingly depend on seamless communication and mutual understanding of actions and intentions. Inspired by Shannon's Information Theory \cite{shannon1948mathematical}, \textbf{synergistic HRC} systems can be viewed as dynamic, bidirectional communication channels in which both agents continuously encode, transmit, and decode information to achieve shared understanding and coordinated action (see Figure \ref{fig:pipeline}).

In this framework, the human acts as a stochastic and context-rich information source, transmitting signals, such as gestures, motions, and haptic or physiological cues, that are intuitive to humans but inherently noisy and high-dimensional. The robot's first challenge is to function as an intelligent decoder (or ``intuitive bridge") that interprets this uncertain stream to infer human state and intent. By maximizing the \emph{mutual information}, i.e., the overlap between human signals and internal task models, the robot should reduce ambiguity and enhance responsiveness.

This decoded information feeds into adaptive task planning, where the robot structures the learned skills and continuously adjusts its strategies based on contextual relevance, task constraints, and user state. Here, the system operates as a ``dynamic decision-maker", prioritizing the most informative inputs. 
Role allocation emerges naturally from this process, and based on the information content and reliability of human versus robot inputs, the system dynamically determines who should lead or assist at any given moment, optimizing for fluency and efficiency. When the roles are determined, the robot control abstraction layer bridges high-level decision-making (like task planning and role allocation) with the real-time execution capabilities of the robot, and executes them by its low-level controller.

\begin{marginnote}[]
\entry{Information theory}{mathematical study of data transmission, compression, and measurement of information.}

\entry{Mutual Information}{measure of the overlap in information content between two signals, i.e., the portion of knowledge that is common to both.}
\end{marginnote}

Just as the revised Shannon's model assumes a feedback mechanism to correct errors \cite{wiener1950human}, synergistic HRC relies on intuitive feedback that is robot-generated signals such as motion cues, haptic guidance, or graphical interfaces %expressive behaviors 
that communicate state, intent, or correction back to the human. This makes the system a closed-loop communication channel, where each action modifies the subsequent \emph{information flow}, allowing online mutual adaptation.

Under this view, synergistic HRC does not imply that every stage of robot programming, task planning, role allocation, low-level control, and feedback must be directly interpretable or perceptually intuitive to the human. Rather, synergistic HRC means that when information flows from humans to robots and returns as feedback, it is encoded, processed, and presented in a way that aligns with non-expert human expectations, minimizes cognitive effort, and supports seamless mutual understanding within the task context.

This review surveys recent advances across the full spectrum of HRC, from the human-to-robot and robot-to-human abstraction layers (often referred to as intuitive bridges), to adaptive task planning, dynamic role allocation, and high-level control abstractions. It highlights key methodological challenges at each layer, deliberately focusing on cognitive and decision-making aspects rather than application-specific implementations. The review also outlines emerging research directions aimed at fostering more natural, resilient, and productive human-robot partnerships. While this review focuses on the cognitive and decision-making layers, it does not cover the robot execution (low-level control) layer. For completeness, readers are invited to refer to \cite{siciliano2010robot, hogan1985impedance, albu2007unified} that comprehensively cover low-level control strategies in robotics.

\subsection{Relation to other surveys}
Recent review papers have provided valuable syntheses of progress in intuitive programming, adaptive task planning, and role allocation. Intuitive programming reviews emphasize natural communication modalities that lower the barrier for non-experts to instruct robots \cite{Kuniyoshi1994, billard2008}.
These works highlight strengths in usability and accessibility but often overlook scalability and high-dimensional input. Reviews on adaptive task planning focus on how robots dynamically modify plans based on changing environments or human states \cite{tsarouchi2016human, tantakoun2025llmsplanningmodelerssurvey}.
Their strength lies in flexibility and resilience, yet many approaches depend on narrowly defined contexts and applications. 
\begin{marginnote}[]
\entry{Intuitive programming}{for robotics refers to designing robot programming methods that are easy to understand and use by non-experts.}
\end{marginnote}

Finally, role allocation reviews examine strategies for distributing responsibilities between human and robot agents \cite{fitts1951human, petzoldt2023review}. 
They offer insights into workload balancing and collaboration efficiency, but overlook real-time adaptability and multi-modal human and robot communication costs or uncertainties.

Individually, these reviews reveal both strengths and gaps in the development of collaborative robots. However, when viewed through the lens of information theory and bidirectional communication, it becomes clear that intuitive programming, adaptive task planning, and role allocation are not isolated capabilities but dynamically interlinked processes that together define the cognitive core of synergistic HRC. This review therefore reframes and integrates these domains within a unified perspective, where mutual understanding, intent inference, and role negotiation are continuously shaped by information flow and feedback. By preserving their internal dynamics while connecting them across the human-to-robot and robot-to-human abstraction layers, we aim to provide a comprehensive understanding of how robots can become more intuitive, adaptive, and responsive in real-world settings.

%\newpage

% %%%%%
\section{HUMAN TO ROBOT INTUITIVE BRIDGE}
\label{sec:human_input}
\begin{figure}[t]
    \hspace{0.3cm}
    \includegraphics[width=0.85\textwidth]{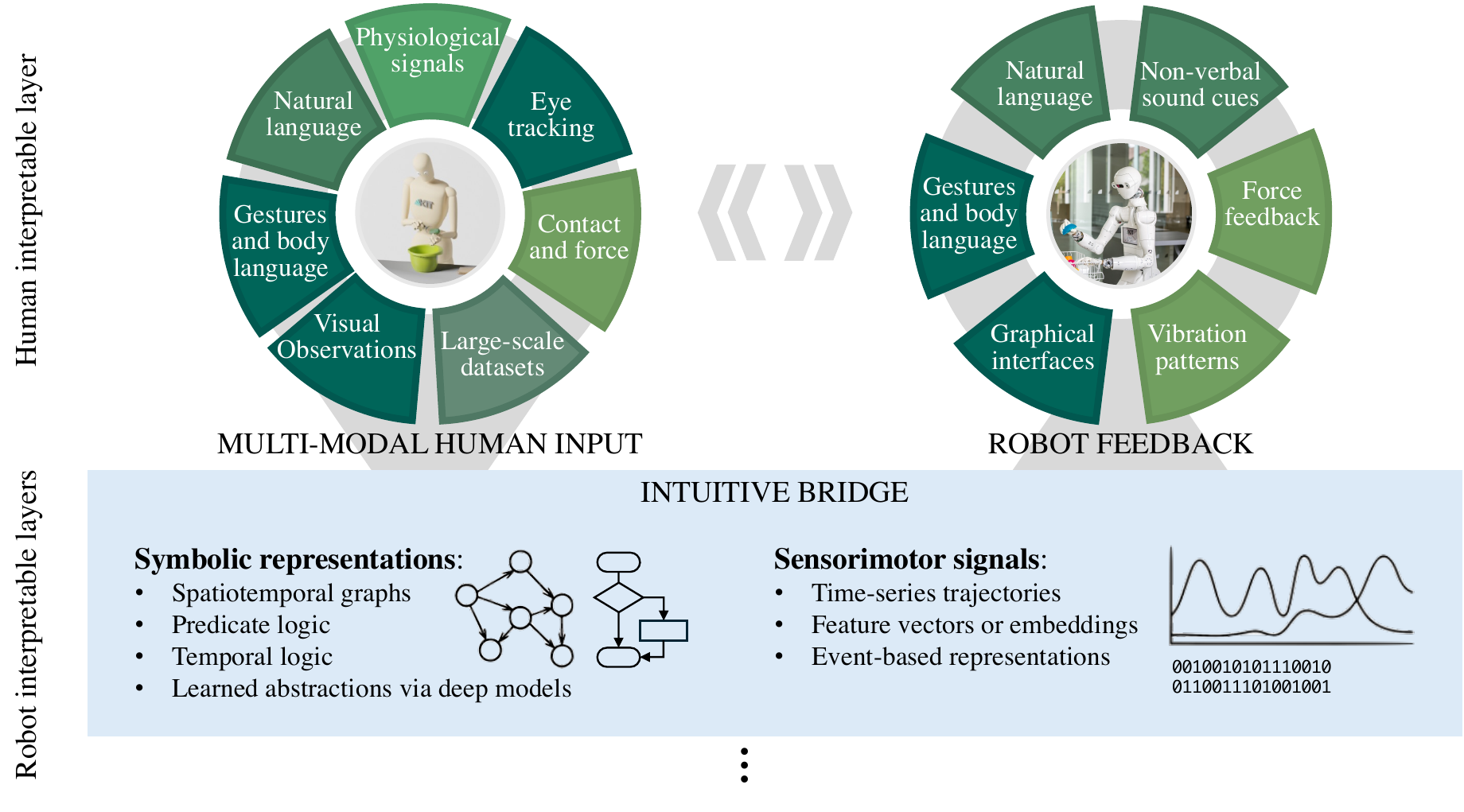}
    \caption{
    The ``intuitive bridge" establishes a bidirectional interface between the human-interpretable layer and the robot-interpretable layers. It functions by translating high-dimensional, multi-modal human inputs into structured symbolic representations
    and sensorimotor signals that robots can process to construct and adapt task models (see Section 2). In the opposite direction, this bridge enables the transformation of robot-generated outputs into human-interpretable feedback channels, fostering mutual understanding between the human and the robot (see Section 6).
    }
    \label{fig:intuitive_bridge}
\end{figure}

% This first section was provided by IIT
%Introduction provided by IIT:\\
This section describes how human state and instructions can be captured through multi-modal inputs and translated into structured, robot-understandable representations. This process constitutes the first stage of the intuitive programming pipeline, establishing one direction of communication within the ``intuitive bridge" (see Figure \ref{fig:intuitive_bridge}), which connects human-interpretable and robot-interpretable layers. The resulting representations should enable online adaptation via this communication bridge, support generalization to novel tasks, and ensure transferability across different robotic platforms.

Traditional robot programming requires specialized knowledge of programming languages, mathematical formulations, and control theory, creating a significant barrier for non-expert users. 
In contrast, the cornerstone of synergistic HRC lies in enabling users to convey intentions, preferences, and knowledge to robots using natural communication modalities that mirror how humans instruct and collaborate. Intuitive robot programming should exploit natural channels (i.e., speech, gesture, demonstration, and even unconscious physiological signals) rather than forcing humans to adapt to machine-readable formats. 

\subsection{Multi-modal Input for Intuitive Programming}
\label{subsec:multi-modal_input}
Human communication is inherently multi-modal, combining verbal instructions with gestural cues, contextual demonstrations, and unconscious physiological responses. Each modality carries distinct information about the task and serves different roles in conveying user intent and preferences.   

\subsubsection{Natural Language} 
Natural language is the most direct and explicit channel for conveying human intent to robotic systems. Through speech and text commands, humans can articulate goals, preferences, and constraints in a familiar format that requires no specialized training or technical expertise. Language excel at conveying abstract concepts, temporal relationships, and conditional logic that might be difficult to demonstrate physically, though grounding linguistic concepts in the robot sensorimotor experience remains challenging~\cite{KruegerEtAlRas2011}.
Recent advances in Large Language Models (LLMs) have dramatically improved robots' ability to interpret language instructions \cite{liang2023code, barmann2024incremental}. However, effective integration requires more than language understanding; it demands translating abstract linguistic inputs into executable robot actions, considering the robot physical capabilities, environmental constraints, and task-specific knowledge. 
Intent extraction from natural language involves the recognition of speaker goals, implicit assumptions, and contextual relevance. 
For instance, when a human says ``be careful with that", the system must infer what ``that" refers to, what constitutes ``careful" behavior in the current context, and how to modify its actions accordingly. This requires sophisticated integration of linguistic interpretation with task reasoning. A robot must disambiguate pronouns and spatial references (``put that over there"), resolve temporal dependencies (``do this after you finish"), and infer missing parameters that humans typically assume from context. For example,~\cite{kartmann2023interactive} presented a method to incrementally learn probability distributions of spatial object relations from few demonstrations and use them to generate object placements based on verbal instructions.

\subsubsection{Gesture and Body Language Based Communication}

Gesture and body language provide powerful channels for communicating spatial relationships, temporal coordination, and online feedback during task execution. Unlike verbal communication, gestural input is grounded in the physical world, making it valuable for tasks involving manipulation, navigation, and spatial reasoning \cite{ferrer2014behavior, bongiovanni2022gestural}.
Hand gestures convey precise spatial information, such as desired object placements or workspace boundaries through pointing, and also express grasping strategies or manipulation preferences through hand shapes \cite{merlo2023automatic}. 
Body language, in turn, provides contextual cues about user attention, confidence, and emotional state, which can inform the robot adaptive behavior \cite{Kulic2005anxiety, lagomarsino2022online, campagna2025estimating}. Deep learning approaches for visual perception have significantly improved the reliability of gesture recognition, but challenges remain in handling variability across users, environmental conditions, and cultural differences.
Dynamic gestures add temporal dimensions to spatial communication, enabling users to indicate sequences of actions, timing constraints, and coordination requirements through continuous motion streams \cite{xu2015online}. For example, a sweeping arm motion might indicate a cleaning trajectory, while rhythmic gestures could specify timing preferences for HRC tasks. 

\subsubsection{Physiological Signals}
Physiological signal processing is an emerging frontier in intuitive robot programming that enables inference of human intent, attention, and cognitive state through biological markers that humans cannot consciously control. This implicit communication channel offers valuable complementary information to explicit verbal and gestural cues. Electromyographic (EMG) signals capture muscle activity and can reveal intended movements before they become visible, enabling proactive robot assistance when combined with force and visual feedback for closed-loop control \cite{ajoudani2012tele, peternel2018robot}.  
Additional signals such as electrocardiography (ECG) and electrodermal activity (EDA) offer insights into psychophysiological arousal and stress during interaction \cite{lagomarsino2025promind, messeri2021human}.
Electroencephalographic (EEG) signals provide information about user cognitive state and attention, though the applications focus on high-level state monitoring due to inherent noise and variability \cite{li2022dynamic}.

\subsubsection{Eye Tracking}
Eye tracking provides valuable insights into human attention patterns, visual focus, and areas of interest that can inform robot behavior and task planning. Gaze patterns reveal not only what humans are looking at but also their search strategies, decision-making processes, and potential intent \cite{%dini2017measurement, 
lagomarsino2022pick, li2022dynamic}.
Integration of such information enables attention-aware adaptation where robots adjust actions based on human gaze patterns \cite{lagomarsino2025promind} or offer additional feedback about objects or regions attracting prolonged attention.

\subsubsection*{Multi-modal Integration}
The future of intuitive robot programming lies in the seamless integration of multiple input modalities that exploit the complementary strengths of natural language, gesture, human demonstrations, and even physiological signals. Advanced fusion techniques must handle temporal synchronization, modal conflicts, and individual user preferences while maintaining online responsiveness and reliability.

\begin{textbox}[b]\section{How can humans intuitively transfer skills to robots?}
Humans naturally learn new skills by observing and mimicking others. Providing \textbf{human demonstrations} is an effective approach for teaching new skills to robots \cite{Kuniyoshi1994, billard_learning_2016}. Demonstrations can be captured through \textbf{multi-modal input channels}, such as visual demonstrations recorded by camera systems, kinesthetic guidance via physical interactions, natural language instructions and explanations, gestural commands and spatial references, and physiological signals reflecting implicit intentions. The key challenge is to derive \textbf{task models} from such multi-modal input, which describe tasks on semantic and sensorimotor level and are executable by and transferable to multiple robots.
\end{textbox}

\subsection{Learning Task Models from Human Demonstrations}

Learning task models that represent information about the task on semantic and sensorimotor levels depends on comprehensive multi-modal capture of human demonstrations, using the input channels described in Section~\ref{subsec:multi-modal_input}, an intuitive way to teach robot new skills (see the sidebar titled How can humans intuitively transfer skills to robots?).
The significant challenge in Programming by Demonstration (PbD) (or Learning from Demonstration (LfD)) is extracting reusable \textit{task models} that represent human motion, objects in the scene, and relations between objects and the human hands \cite{Dreher2020, Dreher2024}.
For the execution on a robot, the learned task models must be mapped to the specific robot embodiment by solving the \textit{correspondence problem}~\cite{argall_survey_2009,billard_learning_2016} to account for different robot physical capabilities, kinematic structures, and sensory configurations, as the demonstrating human and executing robot rarely share identical physical characteristics.

The papers cited in this section were selected through a targeted search combining terms like ``PbD", ``LfD", and ``skill transfer" with keywords related to task models on
\begin{marginnote}[]
\entry{Programming by Demonstration (PbD)}{
fundamental paradigm in intuitive robot programming that enables skill transfer through direct observation and imitation of human actions.
}
\end{marginnote}
 semantic (e.g., ``semantic representation", ``symbolic embeddings", ``learned abstractions") and sensorimotor levels (e.g., ``sensorimotor signals", ``event-based data").

\subsubsection{Learning Constraints from Human Demonstration Data} 
The learned task models must abstract beyond specific instances and generalize to novel tasks while preserving critical constraints on semantic and sensorimotor levels. 
\begin{marginnote}[]
\entry{Task Model}{
a representation of human motion and its variations, capturing spatial, temporal, and physical interaction constraints, along with semantic and sensorimotor information needed for robot execution.}
\end{marginnote}
Key components of effective task models include:
\begin{enumerate*}[label=(\roman*)] 
    \item \textit{Semantic representations} that capture the functional roles of objects, tools, and environmental features involved in the task. This information enables generalization across different objects with similar functions and supports reasoning about task requirements in novel environments.
    \item \textit{Temporal constraints} that specify the required sequencing, synchronization, and timing relationships between different actions and events. These constraints encode when actions must occur in sequence, which can be performed in parallel, and what timing tolerances are acceptable for successful task execution.
    \item \textit{Spatial constraints} that define the relative placement and orientation of objects, tools, and hands in the workspace. These include positional offsets, angular alignments, approach directions, and workspace boundaries that must be respected to ensure safe and feasible task execution.
    \item \textit{Geometric constraints} between keypoints that capture critical spatial relationships such as grasp configurations, contact points, and object-tool interfaces. These constraints often represent the most sensitive aspects of successful task execution.
     \item \textit{Force constraints} that encode physical interaction requirements between the robot, manipulated objects and the environment. Learning force constraints from human demonstrations enables robots to replicate not only the geometric aspects of a task but also the appropriate force application strategies, ensuring safe and effective physical interaction with objects and environments. Such constraints are critical for tasks requiring contact-rich manipulation tasks, precise insertion operations, or collaborative physical interaction with humans.
\end{enumerate*}

Several approaches have been proposed to derive \textbf{symbolic task representations} from demonstrations by integrating semantic knowledge with spatio-temporal constraints. For instance, \cite{ramirezamaro2017transferring} combined object semantics with motion thresholds to classify low-level actions, mapped to robot primitives via a decision tree. %, enabling semantic interpretation of complex tasks. 
\cite{memmesheimer2020robotic} used human pose and spatial object relations to infer actions through ontological reasoning over affordances. \cite{aksoy2011learning} introduced semantic event chains to encode changes of spatial and motion relationships among scene elements over the task execution, while \cite{wachter2013action} used object-action complexes to map observed scene changes to actions. 
\cite{kroemer2016learning} applied random forests to learn objects spatial configurations to fulfill skill preconditions, including object parts and inter-part interactions for improved generalization. \cite{dreher2020learning, wang2023exploring} proposed graph-based methods for action classification and fine-grained interaction recognition, respectively. 
\cite{chou2022learning} encoded skills through linear temporal logic over learned atomic propositions to model action sequences and spatial constraints. %, capturing both the logical sequence of actions and their associated spatial regions, without explicitly modeling low-level motion primitives.
Regarding temporal constraints, \cite{nicolescu2003natural, ekvall2006learning} learned precedence graphs capturing action orderings for sequential tasks, while \cite{dreher2024learning, ye2019robot} also modeled action durations and concurrency, enabling the representation of bimanual activities.

Beyond symbolic abstractions, some methods incorporated \textbf{sensorimotor signals} to retain fidelity of demonstrations while embedding semantic information. In \textit{Kinesthetic Teaching} (KT), robots are physically guided through tasks. Trajectories are recorded via onboard sensors, simplifying the correspondence problem. \cite{steinmetz2019intuitive} combined these trajectories with a continuously updated world model during interaction for skill recognition and parametrization. \cite{zanchettin2023symbolic} introduced semantic networks that capture workspace entities and their relations and vary during demonstrations, allowing the robot to recognize skills' preconditions and effects. \cite{stenmark2018supporting} built an incremental semantic vocabulary from human-robot interactions, enabling robots to enrich their task models while acquiring motion data.
For tasks involving physical interactions, force constraints are embedded alongside geometric data. 
\cite{lambert2019joint} fused vision, tactile, and force-torque data via probabilistic factor graphs, allowing joint estimation of kinematics and forces during manipulation. \cite{lee2020making} encoded visual, depth, force-torque, and kinematic data into a compact latent space using variational Bayesian methods, improving sample efficiency for policy learning in contact-rich manipulation tasks. 

\subsubsection{Policy Learning from Visual Observations} 
\textit{Visual Imitation Learning} (VIL) enables robots to learn from video demonstrations and expands the training data sources to include existing video content and demonstrations performed in different environments~\cite{burke_neural_2010}.
However, it introduces challenges in extracting relevant task information from high-dimensional visual observations and generalizing to novel visual contexts. VIL has been formulated through various computational perspectives:
\begin{enumerate*}[label=\arabic*)]
    \item \emph{Knowledge retrieval}~\cite{pari_surprising_2022,karnan_voila_2022,ramachandruni_attentive_2020}, where the robot retrieves relevant stored demonstrations that align with current perceptual input;
    \item \emph{Motion retargeting}~\cite{qin_dexmv_2022}, which uses optimization techniques to transfer human skeletal motion from videos onto robotic embodiments;
    \item \emph{Image and context translation}~\cite{liu_imitation_2018,sharma_thirdperson_2019,smith_avid_2020}, where implicit visual context from the demonstrator's domain is mapped to the robot context through deep neural networks, bypassing explicit intermediate representations;
    \item \emph{Sequence-to-sequence modeling}~\cite{zhu_learning_2023,fu_humanplus_2024}, where contextual trajectories, such as object and hand motions, are tokenized to train transformer-based policies;
    \item \emph{Constraint learning}~\cite{jin_generalizable_2022,sieb_graphstructured_2019,gao_kvil_2023}, which models explicit spatio-temporal constraints between objects and agents at various representational levels~\cite{kroemer_review_2021}. For instance, \cite{gao_kvil_2023} used statistical methods to extract invariant task features (e.g., keypoints of object functional parts and geometric constraints) from video demonstrations, achieving generalizable, viewpoint-invariant, and embodiment-independent task models.
\end{enumerate*}
Across these formulations, approaches incorporating intermediate or symbolic task representations have shown greater robustness and generalization, enabling robots to adapt skills to novel contexts and variations in task execution.

\emph{Behavior Cloning} (BC) learns a direct mapping from observations to actions without explicitly modeling the underlying task structure. 
While simple, BC suffers from treating observation-action as an isolated input-output pair rather than interpreting the human demonstrated task and the constraints governing it. This makes BC highly sensitive to training data quality and coverage, leading to covariate shift problem when robots encounter states outside the training distribution~\cite{ross_reduction_2011}. 
The black-box nature of BC policies, typically implemented as deep neural networks, provides no interpretable representation of task knowledge, limiting transfer across tasks or embodiments~\cite{chi_diffusion_2023}.  These limitations underscore a fundamental insight: effective imitation learning requires understanding not just what actions humans perform, but how, when and why they perform them. This motivated approaches that model the cognitive processes underlying human demonstrations rather than merely replicating observable behaviors such as \emph{Thought Cloning}~\cite{hu_thought_2023}. 
Incorporating symbolic and linguistic cues can enhance generalization and interoperability.

\subsubsection{Learning from Large-Scale Data}

The scalability of intuitive robot programming heavily depends on the availability and diversity of training data in environments, objects, and human behaviors. The emergence of large-scale, heterogeneous datasets presents both unprecedented opportunities and significant technical challenges. The field is shifting from curated datasets toward leveraging vast, unstructured sources available online, such as internet-scale video repositories, motion capture databases, and ``in-the-wild" recordings.

\paragraph{Unstructured Internet-Scale Data} Several methods aim to extract actionable knowledge from unstructured videos and translate them into robot-executable sequences~\cite{yang2015robot}. While intuitive, this approach often struggles with generalization, as robots replicate observed actions without understanding the underlying task objectives. 
To overcome this, \textit{reward function learning} extracts goal-oriented representations from unstructured video~\cite{chen2021learning} instead of rigid action sequences, enabling adaptation across embodiments and contexts.  
Another promising direction is \textit{world model learning} from large-scale video datasets~\cite{zhou2024robodreamer}, which predicts future environment observations and states and understands object affordances, creating internal simulations to support planning and decision-making.

\paragraph{Structured and Semi-Structured Motion Databases}
Carefully constructed motion capture databases provide high-fidelity, annotated human movement data.
These datasets range from single-view RGB recordings to rich, multi-modal collections including motion capture systems, force sensors, gaze tracking, and physiological signals.
Many incorporate high-level semantic annotations of human demonstrations. Temporal segmentation into atomic actions supports task primitives identification; spatial labeling of objects and body parts enables semantic reasoning about task roles, and natural language descriptions clarify goals and constraints.
Large-scale motion databases such as the KIT Whole-Body Human Motion Database \cite{Mandery2016b} or the Archive of Motion Capture as Surface Shapes \cite{AMASS:ICCV:2019} aggregate multiple sources to create comprehensive repositories of human movement and to enable training of models that can generalize across  tasks, users, and environmental conditions.

\subsubsection*{Technical Challenges}
The integration of heterogeneous data poses challenges: 
(i) \textit{data representation consistency} requires unified synchronization and alignment of diverse sensor modalities; 
(ii) \textit{motion retargeting} becomes complex with large datasets spanning diverse human morphologies and movement styles; 
(iii) \textit{domain adaptation} between data collection and deployment is critical for practical application. Robust intuitive programming systems must account for domain shift while preserving core task knowledge from demonstrations.

\subsubsection*{Opportunities for Scalable and Generalizable PbD}
Given the availability of extensive large-scale data, building \textit{foundation models}, analogous to those in vision and language domains, could be a promising direction for intuitive programming. Current methods target specific applications such as motion retargeting~\cite{yan2023imitationnet} or human-object interactions~\cite{hassan2023synthesizing}, rather than developing general-purpose models for diverse robotic tasks. While the Open X-Embodiment dataset~\cite{o2024open} enables cross-platform training, the resulting models are still tailored to specific robots, tasks, or environments.
Crucially, human motion data encodes expert knowledge and efficient strategies refined through extensive practice, offering unique value beyond what can be acquired through direct robot training.

%%%%%
\section{ROBOT PLANS FOR ADAPTIVE TASK EXECUTION}
\label{sec:task_planning}
This section shifts the focus to task planning, exploring how robots exploit the learned task models to generate executable plans that replicate observed human demonstrations or generalize to novel and dynamic environments while accounting for offline-learned constraints and the continuous information exchange in the human-robot synergy. 
While various models have been developed to represent robotic actions and their effects, the unpredictability of human behavior renders many multi-agent planning approaches unsuitable for HRC. Effective task planning demands adaptive, human-aware systems capable of dynamically updating plans based on a \emph{continuous flow of perceptual and interaction information}. 

This section provides an overview of existing methods for solving dynamic task planning for robots equipped with learned skills.
In particular, we focused on keywords like ``task planning for HRC", ``human-aware task planning", and ``human-in-the-loop robot learning".

\subsection{Single-layer Structures for Task Planning}
\label{subsec:single-layer}

Single-layer task planning approaches aim to formally represent the transition between states and enable reasoning about action preconditions and effects. 

A common symbolic planning formalism is Planning Domain Definition Language (PDDL)  \cite{aeronautiques1998pddl}, which can abstract complex environments into well-defined predicates and actions used to generate deterministic plans for achieving specified goals. 
In \cite{capitanelli2018manipulation}, the authors presented planHRC, a hybrid reactive architecture that uses PDDL to model both the state of an articulated object, i.e., the orientation of each link, and the effects of the actions.
Specifically, when the executed action does not produce the expected effect, the framework asks the human operator to intervene and effectively complete the task. From that point, planHRC allows for the seamless continuation of the task plan.
The work in \cite{andriella2020short} proposed a cognitive architecture for short-term HRC, designed to assist non-expert users during a puzzle game. PDDL was employed both to represent the entire game domain and to plan the robot assistive actions. By monitoring the user's performance and the game progression online, the system selects the action that minimizes a predefined cost function, balancing effectiveness and adaptiveness. 
\begin{marginnote}[]
\entry{Planning Domain Definition Language (PDDL)}{formal language for describing the set of actions, objects, states, and goals, allowing a planner to compute a deterministic sequence of actions from an initial to a goal state.
}
\end{marginnote}%
In \cite{izquierdo2022improved}, the authors proposed a PDDL-based collaborative task planning strategy that explicitly accounts for the human physical and mental state. 
Specifically, the human state was modeled in terms of capacity, knowledge, and motivation, and when a change in this state is detected, the framework performs online replanning.
Implementing the description of the collaborative scenario through PDDL can be difficult, especially when the number of actions and effects grows, leading to increased complexity.

Other approaches were designed to support sequential decision-making under environmental uncertainty, modeling probabilistic mapping from action space to state space.
In HRC scenarios, Markov Decision Processes (MDPs) can represent the human–robot dyad and their interactions, enabling more efficient, human-aware task planning.
In \cite{nikolaidis2015improved}, the authors applied insights from Shared Mental Models (SMMs) and human cross-training practices to enhance HRC planning performance. In particular, MDP was used to represent the team mental model, which the robot exploits to learn preferred human actions and adapt consequently. 
\begin{marginnote}[]
\entry{Markov Decision Process (MDP)}{mathematical framework for modeling decision-making in stochastic environments using states, actions, transition probabilities, and rewards, to compute an optimal plan to reach a goal.
}
\end{marginnote}
The work in \cite{chen2020trust} employed the Partially Observable MDP (POMDP) to allow the robot to infer human trust. This information is then used to adapt the plan to maintain or improve the trust level, maximizing team performance.
In \cite{cramer2021probabilistic}, the authors proposed a POMDP-based planner that adapts to human intentions during a collaborative assembly task. Given the set of sequences to be executed, the robot continuously estimates the human's intended assembly path by observing both part and tool use. Based on this belief, it selects supportive actions aligned with the inferred intent.
MDPs, however, rely on precise models of humans and environments, which are hard to achieve in the real world.

Optimal task planning can be addressed using Genetic Algorithms (GAs) \cite{li2022dynamic} or Mixed-Integer Linear Programming (MILP) \cite{pupa2021human}. Since these methods often also handle allocation and coordination, they are discussed in Section \ref{sec:allocation}.

\subsection{Hierarchical Structures for Task Planning}
\label{sec:hierarchical_planning}
Hierarchical methods organize tasks in layered structures, enabling modular design, and more interpretable decision-making. 
These methods are well-suited for collaborative and long-horizon tasks, where a high-level goal must be decomposed into subgoals.
Given the dynamic nature of HRC, recent research has focused on how these methods can support online adaptation, handling the unpredictability and uncontrollability of human agents.

In \cite{alili2009task}, the authors introduced the Human-Aware Task Planner, a Hierarchical Task Network (HTN)  \cite{georgievski2015htn} planner capable of generating plans that respect social rules.
The work assumed that the human and the robot share the same goal and that the human will follow the plan, thus excluding online adaptation.
In \cite{angleraud2021coordinating}, instead, the authors proposed an architecture to enable a more natural and fluent HRC. The human operator leads the collaborative task and may request assistance from the robot through speech or text commands. The robot responds by analyzing the environment and executing a suitable supportive action using an HTN.
A method for implementing supportive robot behavior is also presented in \cite{mangin2022helpful}. In this work, however, the robot autonomously detects when the human operator needs help and proactively provides assistance. Specifically, robot actions are selected to maximize a desired metric, such as human preferences, by modeling the environment and human state as a POMDP derived from a shared HTN.
The work in \cite{cheng2021human} proposed a framework that leverages human demonstrations.
\begin{marginnote}[]
\entry{Hierarchical Task Network (HTN)}
{planning framework that hierarchically decomposes high-level tasks into simpler subtasks, allowing planners to generate action sequences that respect domain-specific procedural knowledge and constraints.}
\end{marginnote}
Given a set of videos of a human operator performing the task, the system learns the structure of the assembly plan by identifying both sequential and parallel task relationships and automatically constructs a hierarchical task model. At runtime, an optimization-based planner assigns the robot tasks that prioritize actions parallel to those of the human, thereby reducing task completion time and enhancing human satisfaction.
In \cite{ramachandruni2024uhtp}, the User-Aware Hierarchical Task Planning Framework (UHTP), an extension of HTNs for collaborative planning, was proposed. Specifically, whenever the robot is idle, it observes the human operator and generates a set of possible actions, resulting in multiple HTNs. The algorithm selects the plan with the lowest expected cost, adapting to current human behavior.
The limitation of HTNs is the lack of flexibility. HTN planning techniques, indeed, can only generate plans that follow the predefined task decomposition, making adaptation to unexpected events, like unforeseen goals or changes in the environment, difficult to achieve.

An alternative hierarchical task planning structure is the AND/OR graph \cite{de1990and}.
In \cite{johannsmeier2016hierarchical}, the authors proposed a two-layer framework for human-robot task planning and allocation. 
While the first layer generates an optimal task plan using A*, the second layer supports adaptation by locally modifying the graph to handle failures.
The work in \cite{darvish2020hierarchical} introduces FLEXHRC+, a hierarchical architecture for human-robot cooperation. Tasks are represented using a combination of AND/OR graphs and first-order logic, allowing for a compact and modular representation. This formalism supports online decision-making and adapts to variations in human behavior during execution.
\begin{marginnote}[]
\entry{AND/OR Graphs}{
directed structures representing goals or subgoals as nodes, where AND-nodes require all children and OR-nodes only one, capturing conjunctive and disjunctive task relationships.}
\end{marginnote}
In \cite{pupa2022resilient}, the authors proposed a general framework for task planning in HRC scenarios, designed to be resilient to real-world variability. Thanks to the inherent parallelism of AND/OR graphs, the framework builds multiple execution paths that can be selected online to accommodate different human skills and limited resources.
The HATP planning algorithm \cite{alili2009task} is extended in \cite{favier2024model} to enable concurrent execution and respect human decisions. Tasks are encoded as an AND/OR graph where human decisions are modeled as branches. At runtime, the robot generates a policy that selects actions based on partial knowledge of the human's behavior and preferences.
Compared to HTNs, AND/OR graphs offer a more modular representation, supporting local online adaptability. However, they still suffer from structural rigidity. Indeed, when the plan changes or is extended, large parts of the graph often need to be reconstructed. This makes reuse across different scenarios challenging, with limited scalability.

Behavior Trees (BTs) offer greater modularity and reusability. 
In \cite{paxton2017costar}, the authors proposed CoSTAR, a framework designed to let non-expert users create robust robot behaviors through a graphical interface. The system enables the construction of BTs that encode task plans in human-understandable terms and remain reactive to environmental changes. 
In \cite{merlo2024exploiting, merlo2025information} BTs are automatically generated from the analysis of a single human video demonstration of unimanual or bimanual activities using information theory, enabling the replica by one manipulator or a dual-arm robotic system, respectively. Such BTs are then modified according to user preferences expressed in natural language \cite{merlo2025human}. 
The work in \cite{rovida2017extended} combined BTs with HTNs, introducing extended BTs (eBTs). Unlike standard BTs, eBTs incorporate pre- and post-conditions directly into each node. This structure supports more flexible plan optimization and allows for runtime reactivity while preserving a high-level task model.
\begin{marginnote}[]
\entry{Behavior Tree (BT)}{
modular hierarchical architecture used to organize actions and conditions into structured control flows (such as sequences, selectors, and decorators), enabling reactive, flexible, and reusable behavior specification.}
\end{marginnote}
In \cite{fusaro2021human}, the authors applied Utility BTs (UBTs) to HRC scenarios. UBTs are capable of selecting online tasks that minimize a cost function: this feature is used to optimize online a desired collaborative metric, e.g., duration, ergonomics, and travel distance.
Although they are reactive and modular, they can only explore predefined actions, which makes them challenging to use in highly complex and unpredictable scenarios.

\subsection{Learning-based Task Planning}
\label{subsec:learning_based_planning}
Deep learning and Foundation Models (FMs) present a powerful opportunity to generalize and adapt robot plans from multi-modal data. 
Owing to their large-scale training, these models exhibit strong common-sense reasoning abilities, allowing robot systems to interpret and act on human instructions. 
Within this domain, three key paradigms have emerged.
In the \emph{FMs as planner} paradigm, the model directly generates actions step-by-step.
\cite{wang2023demo2code} generated reusable manipulation programs by combining visual demonstrations with language instructions processed using an LLM, including control structures such as loops and conditionals. \cite{wake2024gpt} introduced a multi-modal framework to integrate video and language feedback to construct detailed task plans and affordance maps. \cite{wang2024vlm} developed a VLM-based pipeline to convert human demonstration videos into robotic task plans, with automatic code generation and deployment to simulation and physical robots.

In the \emph{FMs with planner} paradigm, the FMs map instructions to symbolic representations for classical planners, such as a PDDL goal, as in \cite{izquierdo2024plancollabnl, chalvatzaki2023learning}.
Although these methods provide partial guarantees of optimality, they lack the flexibility required for open-ended HRC scenarios.
\begin{marginnote}[]
\entry{Foundation Models (FMs)}{
large neural networks such as Large Language Models (LLMs), Vision-Language Models (VLMs), and Vision-Language-Action models (VLAs) trained on broad internet-scale data and adaptable to specific tasks through fine-tuning.}
\end{marginnote}
To face this challenge, \cite{birr2024} proposed a hybrid approach that combines the flexibility of language models with the correctness of symbolic planners. The system exploits the tool-calling capabilities of LLMs, integrating symbolic planning with scene exploration and object substitution. However, it suffers from challenges such as suboptimal tool selection by the LLM and limited integration of feedback during plan execution. \cite{izquierdo2025raider} built on this by explicitly incorporating reasoning on tool usage within the LLM. 

A third alternative consists of using \textit{FMs as interactive refinement tool} for modifying pre-existing plans, maximizing the adaptability to human preferences and dynamic environments \cite{zhang2024don}. In \cite{bucker2023latte}, humans provide incremental modifications, such as redefining goals, introducing new constraints, or specifying intermediate waypoints, that the LLM interprets to adjust the robot motion accordingly. In \cite{shi2024yell}, continuous natural language feedback was used to guide high-level decision-making, allowing fine-grained adaptation of low-level actions as the task progresses.
The method in \cite{yu2023language} formulated reward functions directly from language input, which are optimized online to translate high-level instructions into robot actions. \cite{cui2023no} introduced a shared autonomy paradigm where verbal instructions and corrections are continuously converted into joystick-like control signals, enabling intuitive real-time manipulation of robotic motion.
To move beyond local motion corrections during execution, 
\cite{merlo2025human} enabled global control of the full task plan and modifications through language commands, while \cite{zhang2024don} exploited LLM to interpret human physically guided corrections of the robot trajectory and refine future execution based on learned adjustments.

Despite rapid advances, these methods face challenges in interpretability and uncertainty estimation, as FMs often produce overconfident answers even in ambiguous situations (hallucinations) instead of seeking clarification. Addressing these issues and integrating continuous learning with online feedback remains an open research direction.

%%%%%
\section{HUMAN-ROBOT ROLE ALLOCATION}
\label{sec:allocation}
Once the robot skills to operate alongside humans and a task plan to achieve a specific goal have been formulated, the next step is to assign responsibility for each action to either the human or the robot.
Traditionally, role assignment was handled through predetermined and static strategies based on the agents' capabilities. Reflecting the principles of Fitts' List of ``Men Are Better At/Machines Are Better At" \cite{fitts1951human}, humans typically handled tasks requiring judgment, flexibility, or creativity, while robots were assigned repetitive, precise, or dangerous work. However, synergistic HRC goes beyond these static divisions \cite{ajoudani2018progress} by including adaptive and context-aware strategies that account for the dynamic state of the co-workers (see the sidebar titled Why dynamically allocate roles?).

\begin{textbox}[h]\section{Why dynamically allocate roles?}
\textbf{Role Allocation} is the process of assigning roles to either the human or the robot in a collaborative task based on decision criteria.
By online monitoring the agents' states and the environment, it is possible to fluidly redistribute roles during execution, optimizing team performance based on current situational demands \cite{petzoldt2023review, lorenzini2023ergonomic}. This allows load balance between the involved agents and increases the \emph{entropy} of agent-to-task assignment combinations to reduce excessive repetitiveness in human tasks and overload \cite{liu2021coordinating}.
\end{textbox}

This section reviews methods for dynamically assigning tasks and responsibilities in HRC. We cover (i) the allocation of tasks or subtasks to either the human or the robot through optimization or search, and (ii) the moment-to-moment tuning of the control authority and workload via adaptive shared control during physical interaction. 
\begin{marginnote}[]
\entry{Role}{the function, responsibility, and set of actions an agent (human or robot) performs within a collaborative task, including both \textit{what} to execute and \textit{how} to behave (e.g., leader, follower, or assistant).}
\end{marginnote}
We categorize these methods according to the primary decision criteria used for allocation: human physical load (Section \ref{sec:allocation_h_physical}), human cognitive load (Section \ref{sec:allocation_h_cognitive}), 
task complexity (Section \ref{sec:allocation_t_difficulty}), and multi-criteria decision strategies (Section \ref{sec:allocation_multi}). Each category explains the motivation for that criterion, the method used for allocation (clustering the approaches in \textit{rule-based}, \textit{optimization-based}, and \textit{learning-based}), and how that cost influences the allocation strategy.

To focus the literature analysis in this section, we used keyword combinations related to allocation (e.g.,
``role allocation", ``task scheduling", ``task assignment"), dynamic adaptation (e.g., ``dynamic", ``online", ``adaptive"), and HRC (e.g., ``human-robot", ``hybrid teams", ``collaborative robot"), excluding works focusing exclusively on static role allocation.

\subsection{Mitigating Human Physical Load and Fatigue}
\label{sec:allocation_h_physical}
In industry or care, heavy lifting and repetitive tasks can lead to musculoskeletal strain and reduced efficiency. Integrating physical state and ergonomics monitoring into role allocation enables smarter load distribution, enhancing human well-being and system performance.

A classical approach is to exploit explicit ergonomic thresholds or \textit{rule-based} conditions to drive allocation decisions.
\cite{makrini2019task} developed an online task allocation system for assembly, segmenting sequences into elementary tasks. The system accounts for agent capabilities and ergonomics, using a Rapid Entire Body Assessment (REBA)-based evaluator to assess human posture. It iteratively assigns tasks, prioritizing the robot when REBA scores exceed safe thresholds, and reorders tasks to avoid successive high-strain activities.
\cite{calzavara2024achieving} used a Mixed Integer Programming (MIP) model for initial task assignment, minimizing makespan while ensuring task feasibility. During execution, online motion tracking estimates fatigue via energy expenditure. If fatigue exceeds safe limits or workload is unbalanced, tasks are reassigned to the robot and returned to the human upon recovery.

To support complex constraints and enable multi-objective trade-offs (ergonomics versus makespan), 
role allocation was framed as a mathematical \textit{optimization-based} decision problem.
\cite{maderna2020online} focused on collaborative kitting, solving a MILP problem to assign picking tasks to the human or robot. The optimization minimizes a cost function, balancing makespan and ergonomic load (using object height, weight, and worker posture evaluated with REBA) while ensuring feasibility and correct sequencing. The plan is updated after each human-completed picking.
\cite{pupa2021human} formulated a MILP for task allocation, optimizing parallelism, precedence, and job quality metrics (e.g., weight, noise, posture). Their two-layer architecture includes a task assignment layer solving the MILP, and a dynamic scheduler that adapts the plan online based on human execution pace, affecting cumulative job quality. The MILP is solved at the end of each job cycle with updated parameters.
\cite{merlo2023ergonomic} proposed an AO$^\star$ search over an AND/OR graph of feasible assembly sequences and agent assignments, with nodes as subassemblies and agent-specific edges as actions. Task costs reflect agent suitability: robots have fixed costs and penalized infeasible tasks, while human costs depend on KWear, derived from Rapid Upper Limb Assessment (RULA) scores and posture duration. When KWear exceeds safe limits, tasks shift to the robot until recovery. After each action, the AO$^\star$ search is re-executed on an updated graph with completed actions removed.

The work in \cite{messeri2022dynamic} adopted a \textit{data-driven} strategy using a 3D vision system and a neural network trained on OpenSim simulations to estimate muscle activations. 
Task selection minimizes alignment between predicted and accumulated fatigue modeled as a 6D joint-based vector, assigning lower-fatigue tasks to the human and demanding ones to the robot.

When tasks cannot be decomposed into subtasks, role allocation may occur at the control level, with the robot adapting its behavior online based on human performance. 
\cite{peternel2018robot} proposed a method where the robot shifts roles in response to motor fatigue, measured via shoulder EMG signals, initially following and gradually taking on more effort as fatigue increases. Building on this, \cite{vianello2024effects} explored human perception and adaptation to abrupt transitions between robot control modes (leader, follower, reciprocal) in a collaborative sawing task.
Although these methods are powerful, they are often tailored to specific tasks and are not easily generalized.

\subsection{Balancing Human Cognitive Load and Preferences} 
\label{sec:allocation_h_cognitive}
A role allocation pipeline that disregards individual cognitive load and personal preferences can compromise overall effectiveness and user satisfaction \cite{lagomarsino2022online}. 
Excessive robot authority may leave the human disinterested or ``out-of-the-loop", whereas insufficient support can lead to cognitive overload. 
Likewise, trust must be carefully calibrated to ensure that the human neither overestimates nor underutilizes the robot capabilities \cite{campagna2024promoting}. 

% Rule-based: if-else, MDP, POMDP
Initial approaches to addressing these challenges relied on \textit{rule-based} logic to assign tasks or switch control based on human mental state thresholds or predefined triggers. 
\cite{giele2015dynamic} monitored the user's cognitive task load and adjusted the robot level of autonomy accordingly. Their method involved dividing the overall task into subtasks, thus offloading work when the human approached overload and handing back tasks when the human could handle more. 
\cite{dubois2020adaptive} instead formulated the task allocation problem as a MDP. At each iteration, their approach determined the fraction of tasks to be carried out by the human, assuming that the user trust level at the previous time step was perfectly observable and including a workload model to inform allocation decisions. 
To account for uncertainty in cognitive state and intention estimation, \cite{roncone2017transparent} employed a POMDP to decide whether to take over a subtask or wait based on a metric such as current human state or minimal completion time, as demonstrated in their experiments. 
By planning in belief-space, the robot could ask clarifying questions or provide information, treating communication as an action to reduce uncertainty. 
\cite{wang2023robot} exploited an online estimate of robot trust and self-confidence to modulate its leader-follower role during physical human-robot interaction (pHRI).

% optimization
Other approaches treated role allocation as an \textit{optimization} problem, often solved with specialized algorithms or control strategies rather than full sequential decision policies. These methods defined an objective function (such as maximizing productivity while keeping stress below a limit \cite{lagomarsino2025promind}) and solved for the best role assignment given the current human state. 
\cite{rahman2018mutual} introduced a two-level feedforward optimization for HRC.
In their setup, the task was pre-divided into subtasks and an offline optimization allocated them based on agents capabilities. During task execution, mutual trust was monitored, and if either the human's trust in the robot or the robot trust in the human dipped below a threshold, the system reallocated the subtask on the fly. 

% -> toward personalisation...
% Learning-based: game-theory, LLM for preferences
Recent approaches have focused on \textit{learning-based} methods that allow the robot to personalize and improve role allocations through data. An example is \cite{ali2022heterogeneous}, who developed a model of robotic trust in the human partner by learning human capabilities over time through observation of task outcomes. Using this learned trust model, the system allocated each new task to the agent expected to maximize the overall reward. 
\cite{messeri2021human} used game theory to model HRC as a non-cooperative game between two players: the robot aimed to maximise productivity (cycle time), while the human sought to minimize stress (estimated by ECG). Each player can choose to adapt (meeting the other player's goal by assuming a follower role) or not adapt (assuming a leader role) to the other. A Learning Automaton then evaluated how the robot role influenced the worker's stress and performance, rewarding role choices that approached the Nash equilibrium.
\cite{izquierdo2024plancollabnl} exploited LLMs to integrate expressed preferences or conditions into task planning. In their system, the human could express their state or desire (e.g., ``I have a headache"), and the LLM translated this into plan adaptations, such as adding specific subgoals and increasing the allocation cost of any task that would aggravate the human condition.

\subsection{Accounting for Task and Layout Complexity and Efficiency}
\label{sec:allocation_t_difficulty}

Task complexity (e.g., number of components and required tools) and environmental constraints (e.g., workspace layout, reachability, safety zones) can intuitively determine whether a human or a robot is better suited for a given job \cite{lagomarsino2022pick}. 

% rule-based
\cite{ranz2017capability} developed a \textit{rule-based} allocation framework that assigned the ``variably distributable" tasks by evaluating their characteristics through multiple criteria and optimizing for process time, quality, and additional investment. Similarly, the work in \cite{malik2019complexity} proposed a complexity-based task classification for assembly, computing time-varying part attributes (i.e., grippability) and the safety/precision demands. High-complexity tasks (e.g., awkward part handling, safety-critical steps) were designated for human execution, while low-complexity, repetitive tasks were automated. \cite{bruno2018dynamic}
combined a hierarchical task decomposition with a capability-informed decision tree classifier, linking each task to possible agents. 
Their rule-based online allocation prioritized loading the robot over the human where possible and supported dynamic reassignments in response to unexpected delays during task execution. 
% graph-based
More recently, \textit{graph-based methods} have been proposed to represent tasks, dependencies, and potential failures or re-routings in a formal structure for role allocation. 
\cite{pupa2022resilient} used an AND/OR task graph to dynamically allocate tasks and reassign them in case of failure or delays. Task and layout complexity are built into this model by including alternative subpaths: for example, if a robot fails at a complex task or encounters an obstacle in the workspace, the graph could switch to a branch where the human performs that task (a resilient fallback).
A graph-based task model was also proposed in \cite{riedelbauch2019exploiting}, where the allocation was defined step by step as the robot continuously evaluated task complexity (in terms of uncertainty in object perception and required actions) and spatial factors (e.g., human proximity), executing tasks only when confident of successful completion.

% optimization
Other approaches integrated cost functions for \textit{optimized search} and path selection over the task graph.
\cite{alirezazadeh2022dynamic} predefined task precedence and estimated execution time, using unexpected delays or speed-ups (measured by observing human progress) as indicators of higher or lower experienced task complexity to trigger rescheduling. A more explicit optimization was developed by \cite{banziger2020optimizing}, who applied a genetic algorithm to adapt role assignments based on task progress, idle time, and travelled distance computed during simulation runs, accounting for potential workspace interference. \cite{petzoldt2022implementation} also proposed an optimization approach to minimize task cycle time. Notably, they implemented their strategy in a no-code, block-based robot planning framework to facilitate intuitiveness and broader adoption. 

\subsection{Multi-Criteria Allocation Strategies}
\label{sec:allocation_multi}
Several criteria can inform allocation decisions to preserve worker health, optimize process efficiency, and ensure task quality. While the above-mentioned approaches typically prioritized a single optimization objective, the works presented in this section combine multiple factors simultaneously. 
\cite{lamon2019capability} proposed a capability-based allocation framework evaluating agent-task pairs by task complexity, agent dexterity (reachability, manipulability, proximity), and physical effort (force, repetition, load). However, optimizing across all criteria is difficult due to trade-offs among competing objectives.
\cite{li2022dynamic} introduced a three-phase dynamic allocation strategy combining EEG and eye-state monitoring, neural network-based fatigue modeling, and reallocation via a multi-objective GA (NSGA-II). After each job batch, human state is updated, and NSGA-II is run to optimize the task sequence and assignment for time, load balance, and satisfaction. Solutions, encoded as binary strings, evolve through crossover and mutation, yielding Pareto-optimal solutions.
In contrast, \cite{lippi2023task} used MILP for task scheduling, optimizing makespan, workload, and quality considering exclusivity, preferences, quality thresholds, spatial conflicts, and switching penalties. The system monitors progress and reallocates pending tasks in response to deviations like performance drops or preference changes. However, MILP high computational cost limits its online functioning.
To mitigate this issue, \cite{fusaro2021integrated} used BTs to structure jobs as task compositions with temporal and logical constraints, enabling decomposition into simpler online sub-problems. A Role Allocator node selects from unlocked tasks and solves a MILP at each trigger, minimizing a fitness function based on agent capabilities, ergonomic impact, and availability. \cite{lamon2023unified} extended this to collaborative settings by adding a suitability cost (assessing agent team compatibility) and a dynamic preference cost that increases with repeated human task rejections. An AR interface facilitates task negotiation, with rejections influencing future allocations.

Most recently, \cite{dimitropoulos2025generative} investigated the use of LMMs to automate task modeling and allocation. By processing raw inputs such as narrated assembly videos and incorporating ergonomic and efficiency prompts, LMMs can generate task sequences and assignments without manual programming. This approach offers high flexibility and adaptability to new instructions or environments, leveraging the power of large pre-trained models.

\begin{figure}[t]
    \centering
    \includegraphics[width=1\linewidth]{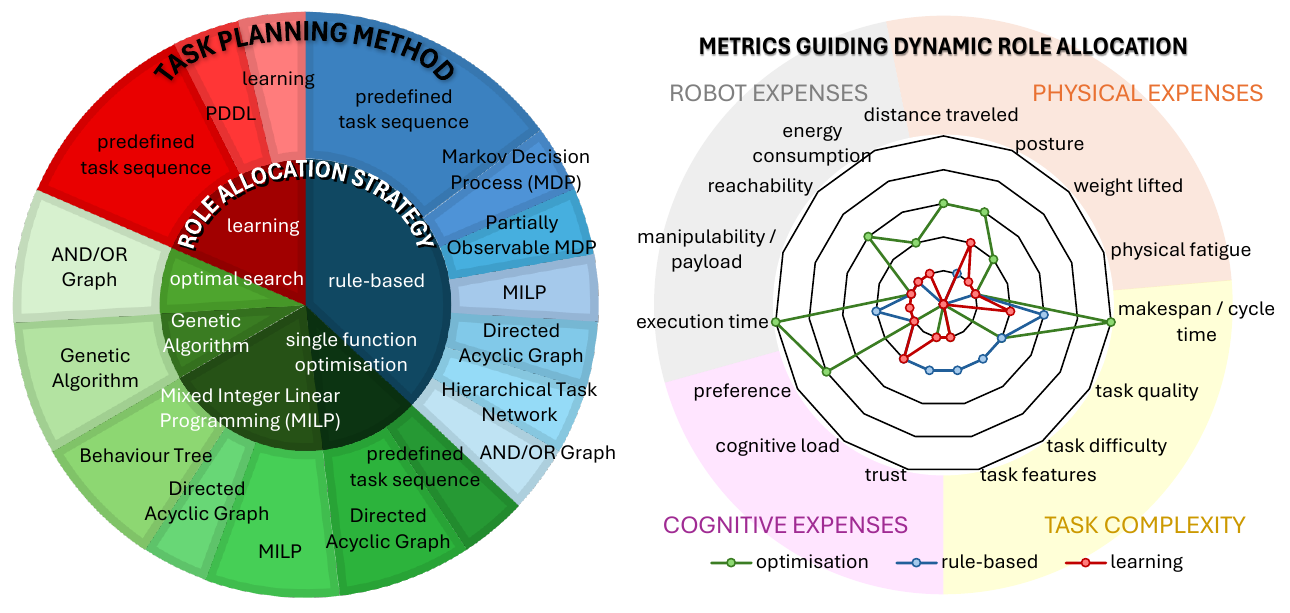}
    \caption{\textit{Left}: Distribution of papers in the literature addressing dynamic role allocation, categorized by the strategy employed (rule-based, optimization, or learning) and their associated task planning methods.
    \textit{Right}: Overview of the papers included in this review, grouped by the dynamic role allocation strategy and the cost guiding the allocation.}
    \label{fig:RA-TP}
\end{figure}

\medskip

Figure \ref{fig:RA-TP}-left presents a dual-ring diagram illustrating the distribution of reviewed papers addressing dynamic role allocation. The inner ring categorizes the allocation strategy into rule-based (dark blue), optimization-based (various green shades), and learning-based (bordeaux). 
Rule-based approaches rely on explicit, human-understandable rules that support transparent decision-making, but their fixed and predefined logic may not fully capture the complexity of collaborative scenarios.
Optimization-based methods offer greater flexibility, yet still depend on predefined objective functions that may limit generalization across users and require high computational complexity. 
Learning-based strategies, though less common, are gaining traction for their ability to personalize and improve role allocation through data-driven adaptation.
The outer ring details the corresponding task planning methods used within each strategy, with segment sizes proportional to the number of papers.
Figure \ref{fig:RA-TP}-right visualizes the cost metrics used to guide role assignment across strategies. 
Colored dots denote the frequency with which each cost appears in the literature per strategy.

\subsubsection*{Holistic Perspective on Dynamic Role Allocation}
Existing dynamic role allocation approaches mostly use human-centered criteria related to workload and preferences. New frontiers extend these models to address the needs of impaired or disabled individuals, thereby improving inclusivity and adaptability in collaborative settings \cite{torielli2024laser, lagomarsino2025mitigating}.
For truly synergistic HRC, however, it is essential to consider the capabilities and constraints of both agents. To this end, emerging frameworks integrate robot-related variables such as reachability, manipulation capability \cite{makrini2019task}, and power consumption \cite{lagomarsino2023maximising} to support more balanced decision-making. 
An open challenge is finding a balanced definition for the costs to effectively distribute the roles in a way that benefits both agents, as humans would do in social contexts.

%%%%%
\section{ROBOT CONTROL ABSTRACTION LAYER} 
\label{sec:control}
Task planning and allocation alone are not sufficient for efficient and reliable HRC. Once a task is assigned, an intermediate layer is needed to translate it into low-level control objectives for the motion controller.
For instance, \cite{pupa2022resilient} mapped each task to a sequence of ordered actions, while \cite{faroni2023optimal} computed an optimal sequence of elementary actions and translates them into low-level commands. This process of converting abstract inputs into geometric-level instructions is known as the symbolic grounding problem \cite{garrett2021integrated}.

During execution, the low-level controller ensures safety by enforcing behaviors that prevent harm to humans.
Several control frameworks explicitly address SSM and/or PFL constraints. In \cite{byner2019dynamic, pupa2025enhancing}, the SSM distance was translated into a velocity constraint enforced at the control level, while \cite{tonola2025reactive} addressed it during planning. For PFL, strategies include modulating velocity based on path deviations \cite{aivaliotis2019power} or using energy tanks to limit system energy \cite{benzi2022energy}. Since focusing on a single paradigm can lead to conservative behavior, some works combined SSM and PFL to enforce more flexible and effective constraints \cite{lucci2020combining}.

Reliable HRC requires more than safety: the robot must adapt to users and contexts, with low-level control considering environment, human needs, and motion predictability.
Compliant and stable behavior is essential for pHRI. Well-established techniques such as admittance control and impedance control \cite{siciliano2010robot, hogan1985impedance} are commonly used. Since impedance control performs well in stiff environments but struggles in soft ones, while admittance control behaves oppositely, \cite{ott2015hybrid} proposed a hybrid controller that continuously switches between the two, leveraging the strengths of both.
\begin{marginnote}[]
\entry{Safety}{in HRC is defined in ISO 10218-1/2 \cite{ISO10218} and ISO/TS 15066 \cite{ISO15066}, outlining four collaborative paradigms: \textit{Safety-rated Monitored Stop} (SMS), \textit{Hand Guiding} (HG), \textit{Speed and Separation Monitoring} (SSM), and \textit{Power and Force Limiting} (PFL).}
\end{marginnote}
Alternative approaches rely on passivity-based control. Energy tanks \cite{benzi2022energy} can stabilize variable admittance controllers in pHRI tasks, including interactions with stiff human operators \cite{ferraguti2019variable} and path-constrained guidance \cite{shahriari2024path}. Admittance controllers can also estimate human intent during collaborative transportation, improving robot adaptability \cite{hamad2021adaptive}, while \cite{yu2021adaptive} used an adaptive impedance controller with neural networks to handle uncertainties.
\cite{haddadin2024unified} introduced the Unified Force-Impedance Control framework, which integrates force and impedance control within a passivity-based approach, leveraging energy tanks to effectively handle uncertainties, contact loss, and controller switching.

To enable safer and more effective collaboration, robot controllers must also account for human-centered factors.
\cite{kim2021human} proposed a method that prioritizes physical ergonomics by estimating joint overloading torques through a whole-body dynamic model and minimizing them online via optimization. \cite{falerni2024framework} used a preference-based optimization, including user feedback and the RULA index, to optimize the robot end-effector pose in object-handling tasks.
Adaptations to user impairments have also been proposed to facilitate the interaction and mitigate compensatory movements \cite{lagomarsino2025mitigating}.
Furthermore, cognitive factors play a key role. \cite{messeri2021human} modelled the interaction as a non-cooperative game, where a learning agent reduces the robot pace online, resulting in decreased human heart rate variability. 
\cite{lagomarsino2025promind} introduced PRO-MIND, a framework that adapts online robot trajectories and safety zones using human attention and psychophysical stress to promote comfort and tune safety limits.

Finally, unpredictable robot motions may compromise safety, potentially triggering involuntary human reactions. 
To mitigate this, \cite{kirschner2022expectable} presented the Expectable Motion Unit, which adapts the robot velocity online to reduce such risks. In \cite{pupa2023dynamic}, predictability was framed in terms of trajectory deviations, with allowable deviation margins adjusted based on operator expertise. The control inputs are computed online to keep the robot trajectory within a tube, whose radius reflects the operator's skill level.

%\newpage

%%%%%
\section{ROBOT TO HUMAN  INTUITIVE BRIDGE} 
\label{sec:robot_output}
Feedback modalities in HRC typically rely on three of the five senses, namely vision, hearing, and touch, while excluding smell and taste. Visual feedback may involve graphical user interfaces, LED indicators, or augmented reality overlays; auditory feedback can range from simple tones to natural language instructions; and tactile feedback includes modalities such as force feedback and haptic vibrations that provide physical sensation or resistance (see Figure \ref{fig:intuitive_bridge}-left). These, however, can be significantly different across important criteria such as \textit{latency}, \textit{multidimensionality}, \textit{precision}, \textit{cognitive load}, \textit{directionality}, \textit{trainability}, and finally, \textit{intuitiveness}. \textit{Latency} describes the speed at which sensory feedback is perceived and acted upon. Auditory and tactile senses typically offer the fastest response times, allowing near-instantaneous perception and reaction. 
Vision follows closely, though it can involve slightly more processing. 
\textit{Multidimensionality} is about how much information a modality can convey at once. Vision performs well here, allowing parallel transmission of spatial, color, and motion data.
Audition also supports high bandwidth, especially in tonal and temporal dimensions.
Touch provides moderate multidimensional feedback (e.g., pressure, texture, vibration). 
\textit{Precision} reflects how finely a sense can detect and differentiate stimuli. Vision and audition again lead, capable of detecting differences in color, shape, pitch, or volume.
Touch can be precise in localized regions (e.g., fingertips), but less so across the body.
\textit{Cognitive load} refers to how much mental effort is needed to interpret the feedback \cite{sweller1988cognitive}. Visual and auditory cues can be low-load if well-designed but become problematic with overload or poor contrast. Haptic feedback often requires less cognitive effort for basic cues. 
\textit{Directionality} is the sense's ability to help localize the source of the feedback. Vision is strongest here, with precise spatial resolution. 
Audition also supports good directionality, especially with binaural cues. 
Touch can indicate source position well if spatially mapped on the body.
\textit{Trainability} concerns how much learning is needed to effectively use the sense for feedback. Vision and audition usually require minimal training, while tactile and proprioceptive cues may require moderate training.

\textit{Intuitiveness} refers to how naturally and effortlessly a sense conveys meaning without requiring conscious interpretation. It tends to be high when a large proportion of the above factors align with human expectations and sensory processing patterns. Indeed, when it comes to HRC applications, visual and auditory feedback tend to be the most intuitive, as we are highly familiar with processing sights and sounds in daily life.
Tactile and proprioceptive cues can also be intuitive when mimicking real-world sensations but remain weaker in comparison to visual and auditory feedback.

The selection of uni- or multi-modal feedback strategies in HRC should be based on balancing these criteria to match the task context and user needs.
To evaluate this choice from an \emph{information-theoretic} perspective, the intuitiveness of a feedback channel can be seen as maximizing the \emph{mutual information}, i.e., the overlap, between the robot internal state (or future actions) and what the human perceives, while also minimizing the uncertainty in the human perception of the robot actual state.
The objective of this section was to close the loop in the human-robot-human information flow by providing a structured definition and analysis of feedback modalities. While several comprehensive reviews already address this topic, focusing on sensory channels \cite{hart2023review}, cognitive mechanisms \cite{herweg2019neural}, and multi-modal integration \cite{zhang2024design}, we synthesized key criteria for effective feedback interaction design. 

%%%%%
\section{DISCUSSION AND CONCLUSIONS} 
This review reframed HRC through an \emph{information-theoretic} lens, conceptualizing it as a continuous and bidirectional exchange of information between the human and the robot. Within this framework, intuitive programming, adaptive task planning, and role allocation are dynamically interconnected processes that together optimize mutual understanding, responsiveness, and task performance. 
Rather than conducting a systematic review, our goal was to provide a different big picture that synthesizes foundational principles and highlights promising research trends for accessible and synergistic HRC. 
We first examined the multi-modal communication channels (such as language, gestures, physiological signals, and demonstrations) through which humans can convey intent and knowledge to robots. We discussed how these inputs can be encoded into structured, robot-interpretable representations to enable the learning of new task models, particularly via PbD, and highlighted the increasing importance of large and diverse datasets.
Next, we reviewed model-based and learning-based approaches for structuring and replicating learned models into goal-oriented plans. 
We then analyzed dynamic role allocation strategies that adaptively assign responsibilities and subtasks based on human physical and cognitive load, preferences, and environmental constraints. These methods aim to balance agent demands while enhancing collaboration fluency through context-aware agent-to-task distribution. 
Although low-level control policies were not reviewed in depth, we outlined key safety and human-factor considerations crucial to synergistic HRC, including compliance control and ergonomic optimization. 
Finally, we closed the loop in the \emph{information flow} by discussing how robots can communicate internal state and forthcoming actions through multi-modal feedback. We highlighted design criteria that facilitate user acceptance and collaborative effectiveness.

The identified future research directions are listed below. The frontier in HRC is to enable online co-adaptation and co-evolution of the robot-user dyad. Beyond static user profiles, future AI-powered collaborative robots should develop personalized, evolving models of their human partners over time. This includes the online inference of human internal states and preferences, as well as the use of generative AI and its common-sense reasoning to proactively assist, augment human capabilities, and foster synergistic collaboration.

% Summary Points
\begin{summary}[SUMMARY POINTS]
\begin{enumerate}
\item Intuitive programming enables non-experts to instruct robots via natural modalities such as speech, gestures, demonstrations, and physiological signals. The paper highlights the need for structured data representations that maximize \emph{mutual information} and reduce ambiguity in human-to-robot understanding.
\item Task planning can rely on symbolic (e.g., PDDL), hierarchical (e.g., AND/OR graphs, BTs), and learning-based methods (e.g., LLMs), each with strengths and limitations. Hybrid approaches combine adaptability to dynamic environments with model interpretability.
\item By modeling allocation as a rule-based, optimization, or learning problem, proposed frameworks mitigate workload, improve team efficiency, and reduce human task repetitiveness through dynamic agent-to-task assignments. The paper categorizes methods across decision criteria and planning strategies (see Figure \ref{fig:RA-TP}).
\item High-level plans are translated into safe, human-aware control commands. The paper outlines criteria for effectively conveying robot state and intent through multi-modal feedback (visual, auditory, haptic) according to task demands and user needs.
\end{enumerate}
\end{summary}

% Future Issues
\begin{issues}[FUTURE ISSUES]
\begin{enumerate}
\item Future HRC systems require real-time, adaptive frameworks capable of fusing large-scale, heterogeneous data streams while handling context shifts, user variability, and noise. To ensure safe and reliable interaction, these systems must also prioritize algorithmic stability, enabling consistent and trustworthy behavior in dynamic and human centric environments.

\item There is a growing need for structured, semantically rich data representations or robot-interpretable sensorimotor signals that bridge human inputs with robot task models. Such embeddings should facilitate generalization across tasks, domains, and embodiments, supporting scalable learning and efficient knowledge transfer. 

\item Task planning and role allocation methods should prioritize transparency and explainability. Future systems should justify their decisions, adapt based on user feedback, and apply uncertainty-aware reasoning to promote trust, safety, and collaborative fluency.

\item Feedback channels must convey robot state and intent with sufficient informational richness while remaining intuitive and cognitively lightweight. Future research should formalize this trade-off and optimize feedback strategies accordingly.

\item Despite the huge potential of the topics reviewed in this article, their integration and validation in real-world scenarios remain at an early stage. 
To fully unlock the potential of HRC systems across domains, future applications must embed these capabilities into practical, real-world deployments.
\end{enumerate}
\end{issues}

%Disclosure
\section*{DISCLOSURE STATEMENT}
The authors are not aware of any affiliations, memberships, funding, or financial holdings that might be perceived as affecting the objectivity of this review. 
\section*{ACKNOWLEDGMENTS}
This paper was supported by the European Union Horizon Projects TORNADO (Grant GA 101189557) and 
HARIA (Grant GA 101070292) and by the Carl Zeiss Foundation through the JuBot project. The authors also thank J. Gao, A. Meixner, C. Dreher, and S. Rietsch.

\bibliographystyle{template/ar-style3.bst}
\bibliography{bibliography}

\end{document}